\documentclass[11pt, onecolumn]{article}
\usepackage{cite}
\usepackage{amsmath,amssymb,amsfonts}
\usepackage{graphicx, subcaption}
\usepackage{mwe}
\usepackage{textcomp}
\usepackage{xcolor}
\usepackage{algpseudocode}	
\usepackage{algorithm}
\usepackage{amsfonts, amsmath, amssymb, amsxtra}
\usepackage{lettrine}
\usepackage{color}
\usepackage{parskip}
\usepackage{float}
\usepackage[flushleft]{threeparttable}
\usepackage[inline, shortlabels]{enumitem}
\usepackage{enumitem, cleveref}
\usepackage{array, makecell}
\usepackage{multirow}
\usepackage{boldline}
\setcellgapes{3pt}
\usepackage{rotating}
\newcolumntype{L}[1]{>{\raggedright\let\newline\\\arraybackslash\hspace{0pt}}m{#1}}
\newcolumntype{C}[1]{>{\centering\let\newline\\\arraybackslash\hspace{0pt}}m{#1}}
\newcolumntype{R}[1]{>{\raggedleft\let\newline\\\arraybackslash\hspace{0pt}}m{#1}}
\usepackage[margin=0.75in]{geometry}

\algnewcommand{\LeftComment}[1]{\Statex \(\triangleright\) #1}
\def\BibTeX{{\rm B\kern-.05em{\sc i\kern-.025em b}\kern-.08em
    T\kern-.1667em\lower.7ex\hbox{E}\kern-.125emX}}

\title{A Stable Combinatorial Particle Swarm Optimization for Scalable Feature Selection in Gene Expression Data
}

\author{{Hassen Dhrif} 
\thanks{Computer Science, University of Miami, Coral Gables, FL, USA, hassen.dhrif@miami.edu} \\
\and
Luis G. {Sanchez Giraldo} 
\thanks{Computer Science, University of Miami, Coral Gables, FL, USA, lgsanchez@cs.miami.edu} \\
\and
Miroslav Kubat 
\thanks{Electrical and Computer Engineering, University of Miami, Coral Gables, FL, USA, mkubat@miami.edu} \\
\and
Stefan Wuchty
\thanks{Computer Science, University of Miami, Coral Gables, FL, USA, wuchtys@cs.miami.edu}}
\date{}

\begin{document}

\maketitle

\begin{abstract}
Evolutionary computation (EC) algorithms, such as discrete and multi-objective versions of particle swarm optimization (PSO), have been applied to solve the Feature selection (FS) problem, tackling the combinatorial explosion of search spaces that are peppered with local minima. Furthermore, high-dimensional FS problems such as finding a small set of biomarkers to make a diagnostic call add an additional challenge as such methods ability to pick out the most important features must remain unchanged in decision spaces of increasing dimensions and presence of irrelevant features. We developed a combinatorial PSO algorithm, called COMB-PSO, that scales up to high-dimensional gene expression data while still selecting the smallest subsets of genes that allow reliable classification of samples. In particular, COMB-PSO enhances the encoding, speed of convergence, control of divergence and diversity of the conventional PSO algorithm, balancing exploration and exploitation of the search space. Applying our approach on real gene expression data of different cancers, COMB-PSO finds gene sets of smallest size that allow a reliable classification of the underlying disease classes.  
\end{abstract}


\section{Introduction}\label{introduction}

Different versions of PSO algorithms \cite{eberhart1995particle,488968,637339,moore1999application} have been designed to select informative markers from gene expression data \cite{han2017gene,mohamad2011modified, shen2008, chuang2008improved, han2014novel, yang2008hybrid}. A necessary condition for these algorithms to scale up to high dimensional feature selection problems is their stability, defined as the robustness of results when uninformative ({\it i.e.} irrelevant \cite{john1994irrelevant}) features are added. While the above-mentioned PSO methods are capable of selecting subsets of predictive genes for reliable disease sample classification, obtained gene subsets are usually large. Such characteristics are rooted in the propensity of PSO algorithms to lose the diversity of the swarm, leading to premature convergence and leaving many areas of the search space unexplored. To address these problems, different solutions including \emph{Algorithm variants}, such as chaotic PSO \cite{kumar2015feature}, fuzzy PSO \cite{agarwal2017frbpso}, among others \cite{chen2014new,yan2016hybrid,zhang2015feature}, as well as \emph{Algorithm modifications} including constriction coefficient \cite{shi2004particle}, velocity clamping \cite{rini2011particle}, among others \cite{hasanzadeh2013adaptive,babu2014face,mukhopadhyay2010cooperating,lim2014particle,lin2008particle}, have been proposed. Furthermore,  \emph{Algorithm hybridization} techniques, where PSO is combined with other evolutionary computation approaches such as genetic algorithms \cite{ghamisi2015feature}, ant colony optimization \cite{kabir2012new} and differential evolution \cite{khushaba2008combined} have been considered as well. 

As sets of biomarkers that allow a reliable diagnostic call need to be limited in size, we designed a novel \emph{algorithm variant} of PSO with multiple \emph{algorithm modifications}, COMB-PSO, that maximizes  (i) the accuracy of sample classification, (ii) minimizes the underlying set size of selected features, and (iii) maintains stability of the size of feature subsets in the massive presence of uninformative ({\it i.e.} irrelevant) features. As a consequence, we expect that our algorithm scales to datasets with tens of thousands of features and few hundreds samples, still selecting relatively small subsets of informative  ({\it i.e.} relevant) features. In particular, COMB-PSO introduces a new encoding technique that preserves the dynamics of the particles by keeping the continuous PSO particle behavior. Such a modification enhances the diversity and search capabilities of the algorithm. Based on the concepts of position and velocity in continuous PSO, COMB-PSO introduces a novel adaptive function that defines inertia weight and acceleration coefficients allowing a more aggressive transition between exploration and exploitation modes. To reduce search time, COMB-PSO features a new dynamic population strategy, allowing the fast discovery of new global best solutions and a turbulence operator lifting the swarm from a local optimum. Enabling the algorithm to find smallest subsets of features, we also introduce asymmetric particle position boundaries. Such a concept constrains the divergence of the swarm and limits positions to those with fewer set bits. As for multi-objective PSO, we introduce a local leader selection mechanism that encourages particles to frequently follow the same leader particle, preserving the coherence of the swarm. Finally, COMB-PSO maintains diversity in the non-dominated set by introducing a similarity based mechanism, keeping small size non-dominated sets. 

To test our approach, we applied COMB-PSO to synthetic datasets where we controlled the amount of informative features. Furthermore, we investigated the stability of COMB-PSO by extending synthetic datasets to higher-dimensional search spaces while maintaining the same informative subsets. We further tested our approach on three disease specific gene expression data sets with high dimensional decision spaces and limited number of samples. In particular, we found that COMB-PSO significantly outperforms the standard PSO variants and scales up to high-dimensional FS problems.  

\section{Proposed Methods}\label{methods}
\subsection{Diversity, exploration and exploitation abilities of the Swarm}\label{new_encoding_mechanism}
\begin{figure}[!tpb]
	\centering
		\includegraphics[width=0.5\linewidth]{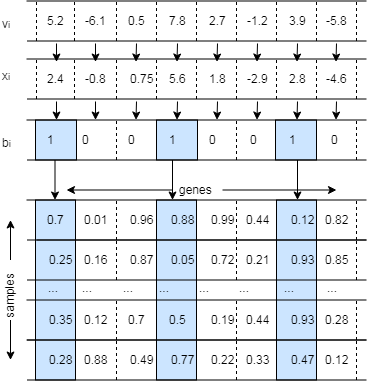}
	\caption{COMB-PSO's encoding mechanism. Aiming at the classification of gene expression profiles, COMB-PSO represents the velocity $v$ and position $x$ of a particle in continuous form. Utilizing a sigmoid function, each dimension of a particle's position is transformed into a binary digit, representing the presence/absence of the corresponding gene in a feature set. As a consequence, only expression values of a gene $i$ with $b_i = 1$ are accounted for in the corresponding classification.}\label{fig:combpso_env}
\end{figure}

As our objective is the selection of a limited set of features, we represent each particle as a binary vector where the presence (absence) of a feature is represented by a binary digit. The binary PSO handles this type of representation by limiting the position of the particles to a binary space where a particle moves by flipping its bits. However, such a movement and the relation between velocity and particle position changes the meaning of velocity. Moreover,  \cite{mohamad2011modified} showed that the sigmoid function, applied to velocity in BPSO, reduces the number of attributes to about half the total number of features only. To circumvent this issue, \cite{mohamad2011modified} introduced a speed component in a modified update rule, where speed controls the probability of a particle changing position. However, such changes hampers the algorithms ability to account for already explored solutions to guide the search. To tackle these issues, we propose a novel encoding scheme (Fig. \ref{fig:combpso_env}) that maps particle positions to probabilities, sustaining search in continuous space. In particular, both velocity $\vec{v}$ and position $\vec{x}$ are represented in continuous form by
\begin{equation}\label{eq:standard_pso_1}
	\vec{v}_{i}(t+1) = \underbrace{\omega  \vec{v}_{i}(t)}_{\text{inertia component}} + \underbrace{r_{1}c_{1}(\vec{p}_{i} - \vec{x}_{i}(t))}_{\text{cognitive component}} + \underbrace{r_{2}c_{2}(\vec{g} - \vec{x}_{i}(t))}_{\text{social component}}
\end{equation}
and 
\begin{equation}\label{eq:standard_pso_2}
	\vec{x}_{i}(t+1) = \vec{x}_{i}(t) + \vec{v}_{i}(t).
\end{equation}
A new binary vector $\vec{b}$ is introduced to map the continuous space position to binary digits by 
\begin{equation}\label{eq:comb_pso:2}
   b_{ij}=\begin{cases}
    1, & \text{if $rand()<S(x_{ij})$}\\
    0, & \text{otherwise}
  \end{cases},
\end{equation}
where \begin{equation}\label{eq:binary_sigmoid}
	 S(x_{ij}) = \frac{1}{(1+e^{-x_{ij}})},
\end{equation}	
indicating that feature $j$ in particle $i$ is accounted for in a feature subset if $b_{ij} = 1$. 

\begin{figure}[!tpb]
	\centering
	\includegraphics[width=0.6\linewidth]{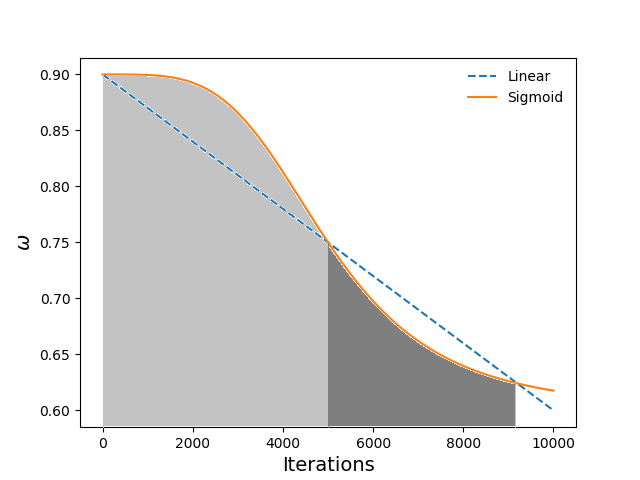}
	\caption{Example of inertia weight as a sigmoid function. Instead of a linear function COMB-PSO applies a sigmoid function to establish inertia weights where, $\omega_{min}=0.6$, $\omega_{max}=0.9$, $a=0.5$ and $b=4$. Compared to a linearly decreasing function, our function maintains longer exploration phase (i.e. shaded area I, where $\omega$ has higher values than linear), as well as longer exploitation phase (i.e. shaded area II, where $\omega$ has lower values than linear). }\label{fig:sigmoid_inertia}
\end{figure}

The terms in Eq. \eqref{eq:standard_pso_1} govern the ability of the swarm to explore the search space. High values of the inertia component, $\omega  \vec{v}_{i}(t)$ and cognitive component, $r_{1}c_{1}(\vec{p}_{i} - \vec{x}_{i}(t))$ in Eq. \eqref{eq:standard_pso_1}, where $p_i$ is the particle specific best solution encountered so far, allow particles to thoroughly {\it explore} the search space. In turn, the social component, $r_{2}c_{2}(\vec{g} - \vec{x}_{i}(t))$, where $g$ is the globally best solution encountered by any particle offsets the effects of the inertia and cognitive components, prompting the particles to converge toward a local optimum. Therefore, dynamics of the particles must be carefully controlled, allowing particles to explore the search space thoroughly in the early stages of a population-based optimization process, without being limited to local optima. In later stages, particles {\it exploit} this information and converge toward the global optimum. Bansal et al.  \cite{bansal2011inertia} compared different inertia weight functions for parameters $\omega, c_1$ and $c_2$ and concluded that despite its popularity the linear time-variant function does not ensure the best performance. Therefore, we propose to use a sigmoid function that allows fast transitions between search phases, extending the particles time to explore and exploit the search space by 
\begin{equation}\label{inertia_weight}
	\begin{aligned}
		\omega = \omega_{min} + (\omega_{max} - \omega_{min}) \frac{1}{1+(\frac{t}{aT})^{b}} \\
		c_{1} = c_{min} + (c_{max} - c_{min}) \frac{1}{1+(\frac{t}{aT})^{b}} \\
		c_{2} = c_{max} + (c_{min} - c_{max}) \frac{1}{1+(\frac{t}{aT})^{b}}.
	\end{aligned}
\end{equation} 
Such a function is shown in Fig. \ref{fig:sigmoid_inertia} where $a$ governs the transition point and $b$ determines the length of the exploration and exploitation phase of the particles. Compared to a linearly decreasing function, our proposed function secures that particles transition fast between full exploration and exploitation modes. We apply the same formula to the cognitive coefficient $c_{1}$ and the inverse formula to the social coefficient $c_{2}$. 

\subsection{Faster Convergence}\label{population_diversity}

To accelerate the convergence process, we introduce a dynamic population strategy. Usually, a new global best solution is discovered at each step while the previous one is discarded. In turn, we keep such solutions and assign the weakest performing particle the corresponding point in the decision space. 

Furthermore, we introduce a turbulence operator: if after $\theta, \theta < N_{iterations}$ consecutive iterations, the global best solution in single objective optimization or the Pareto front in multiobjective optimization did not change, we randomly select a fraction ($\gamma, \gamma \in [0,1]$) of particles that reinitialize their velocities, where $\theta$ and $\gamma$ are user defined parameters. 

\subsection{Smaller Size Subsets}\label{divergence_control}

Particles tend to leave the boundaries of the search space irrespective of the initialization approach \cite{engelbrecht2012particle}. Such behavior mostly translates into wasted search efforts that need to be limited by boundary constraints. However, boundary values of $v_{min}$ and $v_{max}$ have crucial influence as they affect the balance between exploration and exploitation, as well as the size of the generated subsets. If $v_{max}$ is too large, many irrelevant genes will be selected. In turn, some critical features may be missed in the selection process if $v_{min}$ is too small. While most literature adopts symmetric boundaries i.e. $[- v_{max}, v_{max}]$, we introduce an asymmetric boundary coefficient on particle velocities in Eq. \ref{eq:lamda_boundary_coefficient}
\begin{equation}\label{eq:lamda_boundary_coefficient}
	v_{max} = - \lambda v_{min},  \lambda \in [0,1],
\end{equation}
where $\lambda$ is the velocity boundary coefficient. As a consequence, an elevated value of $\lambda$ increases the probability obtaining additional features. 

\subsection{Diversity Preservation of the Pareto Front}\label{dominance_set_maintenance}

\begin{figure}[!htpb]
	\centering
	\includegraphics[width=0.7\linewidth]{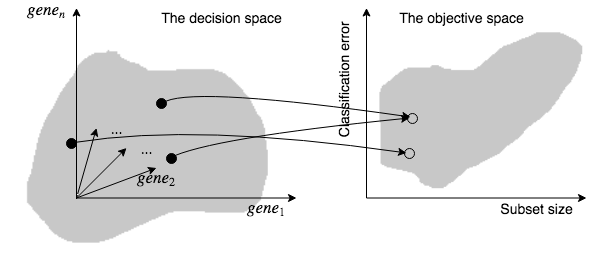}
	\caption{Multi-objective optimization from decision space to objective space}
	\label{fig:map_decision_objective_spaces}
\end{figure}

Several Multi-objective PSO (MOPSO) \cite{coello2007evolutionary} algorithms have been developed based on the Pareto optimality approach that deals with the simultaneous optimization of multiple and conflicting objective functions, without combining them in a weighted sum. The obtained MO solutions are composed of vectors which cannot be improved in any way without causing degradation in at least one of the other objectives. Such solutions are referred to as non-dominated solutions or archive that form the Pareto set (PS) in the search space  (decision space). Its appearance in the objective space that is spanned by the dimensions of the optimization objective, is called the Pareto front (PF). Fig. \ref{fig:map_decision_objective_spaces} depticts the mapping of a decision space into an objective space.  A MOPSO differs from a single objective PSO in that there are multiple global best solutions (called leaders) that each particle is assigned to. The choice of leaders aims at moving the particles towards the PF while ensuring diversity. Usually, all obtained non-dominated solutions are stored in an external archive where leaders are selected from. As a consequence, the main challenge in MOPSO is the suitable selection of a leader to move the particles through the space. In a random approach the selection of leaders \cite{ray2002constrained, mostaghim2003strategies, fieldsend2003using} is stochastic and proportional to certain weights assigned to maintain the population diversity (crowding radius, crowding factor, niche count, etc.). However, such a choice generates a lack of consistency in the movement of the particles and creates a situation of ``craziness'' in the swarm, where one particle follows different leaders at each iteration. In COMB-PSO, each individual particle $\vec{x_{i}}$ picks the leader $\vec{g}_{i}$ that is closest to its own position in the search space instead of the objective space by evaluating Euclidean distances between particles (see Eq.\ref{eq:similarity_distance}).

\begin{equation}\label{eq:similarity_distance}
	\vec{g}_{i} = \arg \min_{\vec{y} \in PS} |\vec{x}_{i} - \vec{y}|.
\end{equation}

To improve the spread of solutions in the Pareto set, PS, particles with shortest distance from their neighbors are removed, usually assessed by the crowding distance (CD) \cite{deb2000fast} and the square root distance (SRD) \cite{leung2014new} measures that are calculated in the objective space. Instead, we use the CD method considering Euclidean distances in the decision space. The particularity of a feature set causes a problem in a multi-objective optimization setting as two particles which are close in the decision space ({\it i.e.} similar selected features) usually induce similar subset size and classification accuracy. However, two particles which are close in the objective space ({\it i.e.} similar classification accuracy and size) might correspond to very different feature subsets. In COMB-PSO, we evaluate dominance based on the objective values; however, in case of equality, a second test is made to further compare both solutions in the search space. If they are not equal, then the decoded solution is added to the PS. 

A last issue with MOPSO, is the choice of control parameters  (inertia factor $\omega$ and acceleration coefficients $c_{1}$ and $c_{2}$)  that govern the convergence behavior of the algorithm to the PF. Chakraborty et al. \cite{chakraborty2011convergence} analyzed the general Pareto-based MOPSO and claimed that if 
\begin{equation}\label{eq:inertia_constraint}
		\omega \leq 1 \quad \text{and} \quad 0 < \frac{c_{1} + c_{2}}{2} < 2(1 + \omega)
\end{equation}
is respected, the MOPSO algorithm will cause the swarm mean to converge to the center of the PS. Our choice of parameters $\omega, c_1$ and $c_2$  satisfy Eq. \eqref{eq:inertia_constraint}, allowing the swarm to converge to the center of the PS as the number of iterations increases. 

\subsection{Problem Representation}\label{objective_function}

In a multi-objective optimization setting, the problem of maximizing classification accuracy and minimizing the set size of features can be formulated in different ways. 
\emph{ The weighted sum method} combines these two objectives into a single objective (SO) using a weight coefficient as follows
\begin{equation}\label{so_fitness_function}
	\begin{aligned}
		\text{Minimize} \quad F(S) = \alpha E_{S} + (1-\alpha) \frac{|S|}{|F|} .\\
	\end{aligned}
\end{equation}
Namely, $S$ is the subset of selected features, while $|F|$ is the total number of features in the whole dataset, and $\alpha$ is a weight factor balancing the number of features and classification performance. Furthermore, $E_{S}$ is the classification error rate. As proposed in \cite{mohamad2011modified}, $\alpha \in [0.6,0.9]$, we set $\alpha=0.8$. Obviously, this problem formulation requires prior knowledge of the search space to choose the appropriate weight parameter $\alpha$. 

Alternatively, the \emph{(b) Pareto front method} formulates the same objective as a multi-objective optimization problem (MO),  

\begin{equation}\label{mo_fitness_function}
	\begin{aligned}
		\text{Minimize} \quad F(S) = [f_{1}(S), f_{2}(S)]\\
		f_{1}(S) = E_{S} \\
		f_{2}(S) = \frac{|S|}{|F|} \\
	\end{aligned},
\end{equation}
representing  a trade-off between classification error rate $E_{S}$ and the number of selected features $|S|$.

\section{Experimental Results}\label{experimental_results}
To test the stability and scalability of our algorithm we created synthetic data sets, establishing a ground truth as to which features are relevant. Furthermore, we obfuscated relevant signals by adding irrelevant features, allowing us to test the algorithms ability to find relevant features in the presence of massive noise. Finally, we applied our algorithms to three cancer related gene expression data sets with a large number of features and a limited amount of samples (Table \ref{tab:datasets}) .

\subsection{Experimental Datasets}
We utilized the last of the three Monks datasets \cite{lichman2013uci} that has 6 discrete features $\{f_{0}, \cdots ,f_{5}\}$. Class labels are 1 if ($f_{3} = 1$ and $f_{4} = 3$) or ($f_{4} \neq 4$ and $f_{1} \neq 3$). Specifically, the Monks dataset has no redundant features, and the most important feature subset is $\{f_{1}, f_{3}, f_{4}\}$. We increased the number of features in the Monks dataset by adding random values, representing irrelevant features \cite{nguyen2018evolutionary}. Likewise, the two other synthetic datasets have 10 continuous features as well. In synthetic 1, features $f_{2}$ and $f_{3}$ are copies of the first two features $(f_{2} = f_{0}, f_{3} =f_{1})$. The class label is set to 1 if the average of the first two features is greater than 0.5. Therefore, there are four optimal feature subsets for this dataset, $\{f_{0}, f_{1}\}, \{f_{0}, f_{3}\}, \{f_{1}, f_{2}\}$ or $\{f_{2}, f_{3}\}$. In Synthetic 2, the first two features are random variables in $[0,1]$. The $3^{rd}$ feature is the average of the first two, $f_{2}=\frac{f_{0}+f_{1}}{2}$ while the $4^{th}$ feature is a copy of the first feature,  $f_{3}=f_{0}$. As a consequence, there is redundancy in any feature subset that contains $f_{0}$ along with $f_{3}$ or ($f_{1}$ and $f_{2}$). The class label is determined by feature $f_{2}$. In particular, the class label is set to 1 if $f_{2} > 0.5$, suggesting that the optimal feature subset in this dataset is $\{f_{2}\}$. In both synthetic 1 and synthetic 2, the rest of the features are irrelevant and have random values in [0,1]. To test the stability and the scalability of the different algorithms, we exponentially expanded the three synthetic datasets up to to $10^{2}$ and $10^{4}$ features by adding noise features (random values in [0,1]). 

\begin{table}[!htb]
      \centering
			\caption{Characteristics of synthetic and gene expression data sets}
			\begin{tabular}{l | c | c | c | c | c}
				\hline
				Dataset & \# samples ($m$) & \multicolumn{3}{c|}{\# features ($n$)}  & \# classes \\
				\hline 
				\multicolumn{6}{c}{SYNTHETIC DATASETS} \\
				\hline
				Monks 			& 432 & 10 & $10^{2}$ & $10^{4}$ & 2 \\
				Synthetic 1 & 200 & 10 &$10^{2}$ & $10^{4}$ & 2 \\
				Synthetic 2 & 200 & 10 & $10^{2}$ & $10^{4}$ & 2 \\
				\hline
				\multicolumn{6}{c}{GENE EXPRESSION DATASETS} \\
				\hline
				Leukemia  & 72 & \multicolumn{3}{c|}{12,582} & 3 \\
				Prostate Tumor & 95 & \multicolumn{3}{c|}{16,535}  & 2 \\
				B-cell lymphoma & 77 & \multicolumn{3}{c|}{6,428}  & 2 \\				
				\hline \hline
			\end{tabular}\label{tab:datasets}
\end{table}
As for real gene expression data sets (Table \ref{tab:datasets}) we used a set of 72 \emph{Leukemia}  patient samples \cite{Armstrong2001}. In particular, this set consisted of 28 Acute Myeloid Leukemia (AML), 24 Acute Lymphoblastic Leukemia (ALL) and 20 Mixed-Lineage Leukemia (MLL) cases, capturing the expression levels of 12,582 genes.  The \emph{Prostate Tumor} \cite{SINGH2002203} data set measured expression levels of 16,535 genes in a total of 95 samples where 52 samples referred to tumor samples and the remainder to non-disease controls.  The \emph{Diffusive Large B-Cell Lymphoma} data set \cite{Shipp2002} captures 58 patients with Diffuse Large B-Cell Lymphomas (DLBCL) and 14 patients with Follicular Lymphomas where each patient sample features 6,428 genes.

\begin{table}[!htb]
			\centering
			\caption{Hyper parameters used in the expreimental set-up. Symbol $\omega$ is the inertia weight, $c_{1}, c_{2}$ the velocity coefficients, $\lambda$ the velocity boundary coefficient (e.g. for dataset of size 10 $\lambda=- v_{max}/v_{min}=1/3$), $\theta$ the number of iterations gbest being trapped before firing the turbulence operator and $\gamma$ the swarm fraction of the turbulence operator (i.e. the percentage of the swarm resetting their velocities)}
			\begin{tabular}{l | c c V{3} c c V{3}  c c}
				\hline
				\multirow{3}{*}{Parameters}  & \multicolumn{6}{c}{\# Features} \\
				\cline{2-7}
				 	& \multicolumn{2}{c V{3}}{10 } & \multicolumn{2}{c V{3}}{100 }  & \multicolumn{2}{c}{$\ge$10000 } \\						 
				\cline{2-7}
				 & min & max & min & max & min & max \\
				 \hline
				\multicolumn{7}{c}{FOR ALL ALGORITHMS} \\
				\hline
				$\omega$ & 0.4 & 0.6 & 0.4 & 0.8 & 0.4 & 1.0 \\
				$c_{1}, c_{2}$ & 1.7 & 2.1 & 1.7 & 2.1 & 1.7 & 2.1 \\
				velocity & -3.0 & 3.0 & -4.0 & 4.0 & -6.0 & 6.0 \\
				swarm size & \multicolumn{2}{c V{3}}{30} & \multicolumn{2}{c V{3}}{100} & \multicolumn{2}{c}{300} \\
				\# iterations & \multicolumn{2}{c V{3}}{300} & \multicolumn{2}{c V{3}}{1000} & \multicolumn{2}{c}{3000} \\
				\hline 
				\multicolumn{7}{c}{FOR COMB-PSO ONLY} \\
				\hline 
				velocity ($\lambda$) & -3.0 & 1.0 (1/3)& -4.0 & 0.5 (1/8)& -6.0 & 0.25 (1/32) \\
				position & -3.0 & 3.0 & -4.0 & 4.0 & -6.0 & 6.0 \\
				$\theta$ & \multicolumn{2}{c V{3}}{5} & \multicolumn{2}{c V{3}}{5} & \multicolumn{2}{c}{5} \\
				$\gamma$ & \multicolumn{2}{c V{3}}{20\%} & \multicolumn{2}{c V{3}}{20\%} & \multicolumn{2}{c}{20\%} \\
				\hline \hline
			\end{tabular}\label{tab:parameters}	
\end{table}

\subsection{Experimental Setup}
\begin{table*}[t]
\scriptsize
		\caption{Performance results using synthetic datasets MONKS, Synthetic 1 and 2. Comparing performance of results that we obtained with the single objective SO-COMB-PSO, the multi-objective MO-COMB-PSO and standard approaches binary BPSO and multi-objective MOPSO, we measured the mean sizes of feature sets $\langle FS \rangle$, mean classification error $\langle E \rangle$, the average percentage cover of strongly relevent features $\langle \% SRF\rangle$  that feature subsets captured as well as the mean number of function calls $\langle FC \rangle$.  }
		\centering
		\begin{tabular}{ l |  c | c | c || c | c | c V{3} c | c | c || c | c | c  }
			\hline
			Mode & \multicolumn{6}{c V{3}}{SINGLE} 	& \multicolumn{6}{c}{MULTIPLE} \\
			\hline			
			Algorithm & \multicolumn{3}{c||}{BPSO} &  \multicolumn{3}{c V{3}}{SO-COMB-PSO} %
			 & \multicolumn{3}{c||}{MOPSO} &  \multicolumn{3}{c}{MO-COMB-PSO} \\
			\hline
			\# features & 10 & $10^{2}$ & $10^{4}$	& 10 & $10^{2}$ & $10^{4}$	& 10 & $10^{2}$ & $10^{4}$	& 10 &$10^{2}$ & $10^{4}$	\\		
			\hline
			\multicolumn{13}{c}{\textbf{MONKS}} \\
			\hline 
			$\langle FS \rangle$ 	& 3 & 15 & 5 056 & 3  & 3 & 25 &  4 & 17 &  4 978 &  4 & 5 &  17\\
			$\langle E \rangle$ 		&  0.0\% & 3.8\% & 5.6\% &  0.0\% & 0.1\% & 5.4\%&  16.0\% &  5.3\%&   7.8\%&  12.5\% & 0.0\% &  3.2\%  \\
			$\langle \% SRF\rangle$  	& 100\% & 50\% &  100\%& 100\% &  100\% &   100\%&  38.5\% &  62\% &  78\%&  44.4\% & 100\% &   56.9\% \\
			$\langle FC\rangle$ 	& 333 & 10 650 &  480 986 &  246 	&  23 280  &  567 300&  173 & 31 120  &  670 980 &  320 & 11 380 & 774 000\\
			\hline
			\multicolumn{13}{c}{\textbf{SYNTHETIC 1}} \\
			\hline 
			$\langle FS \rangle$  	&  2& 31 &   5 209 & 2  & 6 &  249 &  3 & 17 &  4 292 &  4 & 5 &  18\\
			$\langle E \rangle$ 		&  5.2\% & 6.2\% &  5.8\% &  5.0\% & 4.4\% &  13.3\% &  10.9\% &  6.6\%&  30.6\% &  11.7\% & 2.8\% &   10.1\% \\
			$\langle \% SRF\rangle$   	& 100\% & 100\% &  100\%& 100\% &  100\% &  100\% &  85.7\% &  98.4\% &  100\% & 77.4\% & 100\% &  100\%  \\
			$\langle FC\rangle$  	& 276 & 27 170 &   449 876 &  308 &  22 330  &  434 800 &  761 & 24 540  &  342 800&  2 475 & 45 540 & 808 980\\
			\hline
			\multicolumn{13}{c}{\textbf{SYNTHETIC 2}} \\
			\hline 
			$\langle FS \rangle$  	& 1 & 18 &  5 032 & 1  & 1 &  90 &  2 & 15 &  4 459 & 4 & 5&  11\\
			$\langle E \rangle$ 		&  0.0\% & 0.1\% &  5.6\% &  0.3\% & 0.1\% &  0.1\% &  9.8\% &  8.3\%& 20.3\% &  1.9\% & 0.0\% &   6.0\% \\
			$\langle \% SRF\rangle$  	& 100\% & 100\% & 100\% & 100\% &  100\% &  100\% &  72.7\% &  74.1\% &  92.3\% &  83.3\% & 100\% &  86.7\% \\
			$\langle FC\rangle$  	& 623 & 9 470 & 582 225 &  541 &  11 965  &  333 685&  309 & 19 580 & 391 740 &  342 & 8 209 & 397 568\\
			\hline \hline
		\end{tabular}\label{tab:synthetic_results}
\end{table*}

Simulations were carried out with numerical benchmarks to find the best ranges of values. An improved optimal solution for most of the benchmarks was observed when $c_{min} = 1.7$  and $c_{max}=2.1$. As shown in Table \ref{tab:parameters}, however, the boundaries are sensitive to the size of the dataset when it comes to the inertia weight, the velocity, and the positions. Parameters $\omega, c_{min}$ and $c_{max}$ were set with respect to Eq. \ref{eq:inertia_constraint}.

COMB-PSO uses a wrapper approach, requiring a machine learning model to evaluate the classification accuracy of the selected features. Here, we use Random Forest (RF),  an ensemble classification algorithm well suited for microarray data \cite{diaz2006gene}. In particular, we used mtry= $\sqrt{n}$, ntree = 5000, nodesize = 1, where $n$ is the number of features, mtry is the number of input variables tried at each split, ntree is the number of trees in each forest and nodesize is the minimum size of the terminal nodes.

During the search process,  70\% of randomly selected instances were used as training set, while the remaining 30\% were used as test set. Furthermore, we utilized 10-fold cross-validation (on the 70\% training data) to evaluate the classification accuracy of the selected feature subset. To obtain testing classification accuracy selected features were evaluated on the test set (on the 30\% test data).

\subsection{Results}\label{comparison_results}

Based on the choice of the objective functions in section \ref{objective_function}, we investigated the performance of single objective variant SO-COMB-PSO and the multi-objective variant MO-COMB-PSO, comparing their performances to the standard BPSO and MOPSO respectively. Specifically, SO-COMB-PSO and BPSO are single objective methods producing 30 solutions for each dataset from 30 independent runs. MO-COMB-PSO and MOPSO are multi-objective algorithms producing 30 sets of solutions for each dataset from 30 independent runs as well. 

Before introducing the benchmarks used for evaluating and comparing results, we make a quick note on the concept of strongly relevant features. As outlined in \cite{john1994irrelevant},  features are classified into three disjoint categories, namely, \emph{strongly relevant}, \emph{weakly relevant}, and \emph{irrelevant} features. However, simply combining a highly ranked feature with another highly ranked feature often does not form a better feature set because these two features could be highly associated. In fact, each optimal subset must include all the strongly relevant features. It may include some of the weakly relevant features, but none of the irrelevant features. Identifying strongly relevant features is of particular interest to the  high-dimensional FS problems such as finding a small set of biomarkers to make a diagnostic call. Consequently, for the synthetic datasets, where we have prior knowledge of the strongly relevant features, we measured the feature subsets propensity to cover all the strongly relevant features by the  percentage of strongly relevant features that appear in the corresponding feature subsets $\langle \% SRF\rangle$. 

For the rest of the benchmarks allowing us to compare the performance of the algorithms on the synthetic datasets we collect the average size of the optimum feature subsets $\langle FS\rangle$ and the average  classification error rate, $\langle E \rangle$ after 30 runs in Table \ref{tab:synthetic_results}.  Finally, we evaluate the average number of function calls to the classification procedure $\langle FC\rangle$, indicating the algorithm's convergence toward a local minimum. Given that all algorithms appeared to have the same running time when the size of the dataset was fixed, a low $\langle FC\rangle$ may also indicate premature convergence. 

in Table \ref{tab:synthetic_results} we generally observe that our novel variants of COMB-PSO allowed (i) high classification accuracy with (ii) a low size of feature subsets that (iii) largely captured strongly relevant features. Notably, these results were largely independent from the number of available features in the corresponding datasets. In turn, we observed that the conventional PSO variants were sensitive to an increasing number of features. Notably, the average number of calls to the underlying classification function increased sharply with increased number of features, an observation that was independent of the utilized PSO variant. As for single and multi objective specific characteristics our results suggest no significant differences when we compared average subset size, classification error rate, and cover rate of strongly relevant features of SO-COMB-PSO and MO-COMB-PSO. 

While synthetic data sets featured a large numbers of features and samples, our gene expression datasets were high-dimensional as well but had a low number of samples. Applying our algorithms to the three different cancer gene expression data sets, we evaluated their performance by measuring the mean size of of selected gene subsets and the corresponding mean classification error rate.  For the multi-objective approach, the 30 sets of solutions are combined together to extract the overall non-dominated solutions achieved by MO-COMB-PSO and MOPSO across the 30 independent runs. 

\begin{figure*}[ht]
    \begin{subfigure}[t]{0.3\textwidth}
    \centering
    Leukemia2
        \raisebox{-\height}{\includegraphics[width=\textwidth]{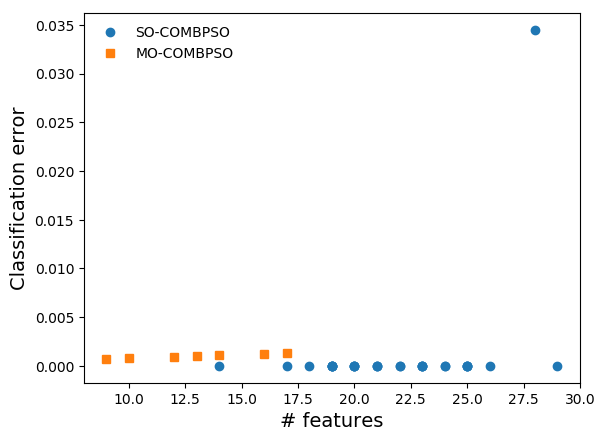}}
        \caption{}
    \end{subfigure}
    \begin{subfigure}[t]{0.3\textwidth}
    \centering
    Prostate tumor
        \raisebox{-\height}{\includegraphics[width=\textwidth]{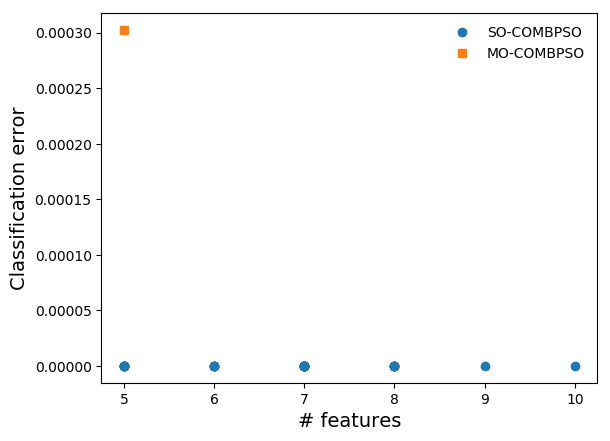}}
        \caption{}
    \end{subfigure}
    \begin{subfigure}[t]{0.3\textwidth}
    \centering
    B-cell lymphoma
        \raisebox{-\height}{\includegraphics[width=\textwidth]{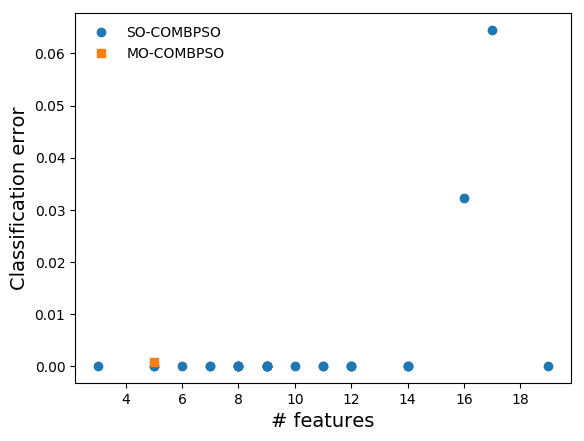}}
        \caption{}
    \end{subfigure}

    \begin{subfigure}[t]{0.3\textwidth}
        \raisebox{-\height}{\includegraphics[width=\textwidth]{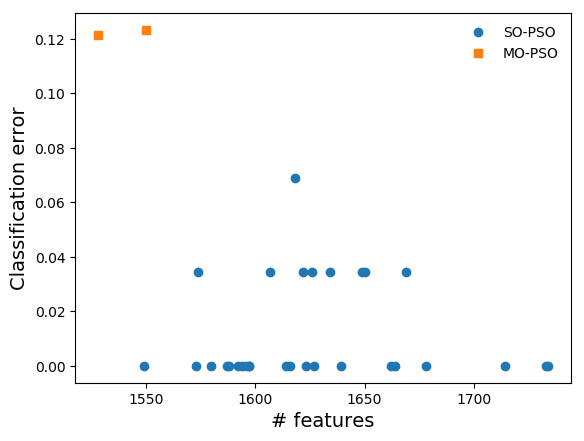}}
        \caption{}
    \end{subfigure}
    \begin{subfigure}[t]{0.3\textwidth}
        \raisebox{-\height}{\includegraphics[width=\textwidth]{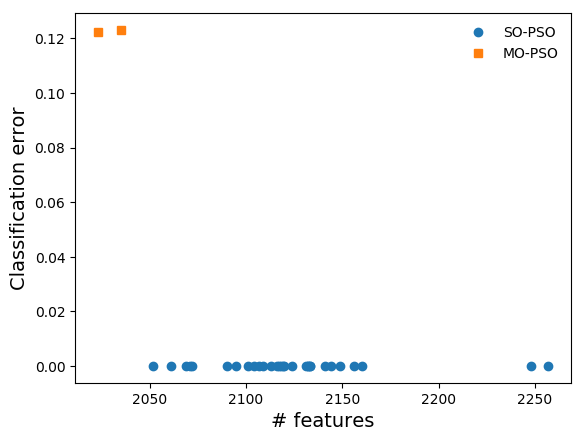}}
        \caption{}
    \end{subfigure}
    \begin{subfigure}[t]{0.3\textwidth}
        \raisebox{-\height}{\includegraphics[width=\textwidth]{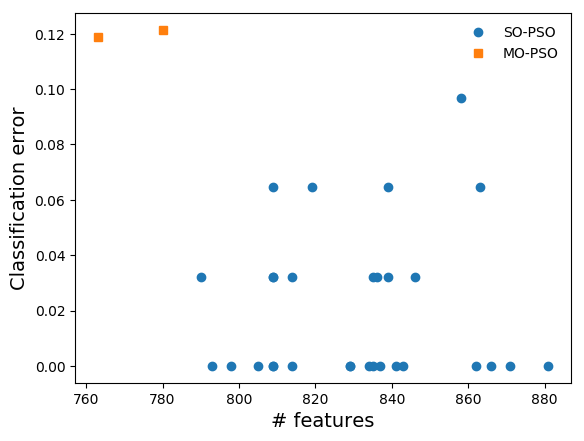}}
        \caption{}
    \end{subfigure}
    \caption{single vs multi-objective and standard PSO ( BPSO, MOPSO) vs COMB-PSO performance on gene expression datasets. Figs a-c show COMB-PSO results, while Figs d-f show BPSO and MOPSO results. Note that for the single objective variants (i.e. BPSO and SO-COMBPSO), different solutions may have the same number of features and the same classification error and they are plotted in the same dot, which explains why some plots have less than 30 dots. Furthermore, we merged the 30 Pareto sets to retain the overall dominant solutions only, which explains why the number of squares is also less than 30.}
		\label{fig:combpso_bpso}
\end{figure*}

The results obtained by BPSO and MOPSO (Figs. \ref{fig:combpso_bpso}d-f) are largely in consent with the  recent ones published on the same datasets in \cite{tran2016investigation}. In comparison to the results that we obtained with the standard PSO variants, we generally observed that our variants of COMB-PSO allowed us to find significantly smaller gene subsets with higher ability to correctly classify disease samples (Figs.  \ref{fig:combpso_bpso}a-c). As for results that we obtained with our variants of COMB-PSO  suggest that MO-COMB-PSO and SO-COMB-PSO generated subsets that roughly allow the same high classification accuracy. However, we found that the multi-objective COMB-PSO outperforms the single objective version by providing smaller gene subsets.  COMB-PSO results are 

\section{Conclusion}
Our objective was to enhance the PSO's performance, allowing us to find feature subsets that improve classification accuracy when implemented on datasets with tens of thousands of features and few hundreds samples. Improving PSO's stability and scalability we introduced (i) a new encoding scheme in the continuous space, (ii) a fast varying inertia weight and acceleration coefficients, (iii) a novel diversity strategy, and (iv) an enhanced local guide search mechanism for the multi-objective approach. Notably, such algorithmic changes enabled us to identify subsets of considerably smaller size and low classification error when we compared their performance to standard variants BPSO and MOPSO.

Comparing the single to the multi-objective approach, we generally observed similar performance in terms of selected subset size, classification error rate, strongly relevant features cover and number of function calls. Nevertheless, the single objective approach can still identify solutions that are not dominated by the best Pareto set of the multi-objective approach. Such an observation may be rooted in the choice of a weight factor in the weighted sum approach, that privileges accuracy over size.  

Since the fitness function deployed in this study does not consider the relevance of the selected subset nor the redundancy of features, our algorithm can not account for  the relevance of genes for the underlying disease class. To capture such characteristics, functional analysis of the underlying genes may be needed to identify clusters of related genes as a proxy to assess their redundnacy as fetaures in the classification step. Furthermore, relevance and redundancy of each feature may be need to be accounted for in the objective optimization function. Although our approache does not explicitely consider redundancy of features, we observed that strongly relevant features were still selected as fetauers in the corresponding feature subsets. While our wrapper method using a random forest  classifier is very efficient at obtaining high accuracy subsets, it still has high computational costs may not scale up to problems with large datasets  Therefore, further research efforst are needed to explore the stability and scalability as well as mitigate the computation costs of the algorithm when dealing with an ultra large dimensional search space, that is composed of millions of features and hundreds of thousands of samples.

\bibliographystyle{IEEEtran}
\bibliography{Bibliography}

\begin{thebibliography}{10}
\providecommand{\url}[1]{#1}
\csname url@samestyle\endcsname
\providecommand{\newblock}{\relax}
\providecommand{\bibinfo}[2]{#2}
\providecommand{\BIBentrySTDinterwordspacing}{\spaceskip=0pt\relax}
\providecommand{\BIBentryALTinterwordstretchfactor}{4}
\providecommand{\BIBentryALTinterwordspacing}{\spaceskip=\fontdimen2\font plus
\BIBentryALTinterwordstretchfactor\fontdimen3\font minus
  \fontdimen4\font\relax}
\providecommand{\BIBforeignlanguage}[2]{{%
\expandafter\ifx\csname l@#1\endcsname\relax
\typeout{** WARNING: IEEEtran.bst: No hyphenation pattern has been}%
\typeout{** loaded for the language `#1'. Using the pattern for}%
\typeout{** the default language instead.}%
\else
\language=\csname l@#1\endcsname
\fi
#2}}
\providecommand{\BIBdecl}{\relax}
\BIBdecl

\bibitem{eberhart1995particle}
R.~Eberhart and J.~Kennedy, ``Particle swarm optimization, proceeding of ieee
  international conference on neural network,'' \emph{Perth, Australia}, pp.
  1942--1948, 1995.

\bibitem{488968}
J.~Kennedy and R.~Eberhart, ``Particle swarm optimization,'' in \emph{Neural
  Networks, 1995. Proceedings., IEEE International Conference on}, vol.~4, Nov
  1995, pp. 1942--1948 vol.4.

\bibitem{637339}
J.~Kennedy and R.~C. Eberhart, ``A discrete binary version of the particle
  swarm algorithm,'' in \emph{1997 IEEE International Conference on Systems,
  Man, and Cybernetics. Computational Cybernetics and Simulation}, vol.~5, Oct
  1997, pp. 4104--4108 vol.5.

\bibitem{moore1999application}
J.~Moore and R.~Chapman, ``Application of particle swarm to multiobjective
  optimization,'' \emph{Department of Computer Science and Software
  Engineering, Auburn University}, vol.~32, 1999.

\bibitem{han2017gene}
F.~Han, C.~Yang, Y.-Q. Wu, J.-S. Zhu, Q.-H. Ling, Y.-Q. Song, and D.-S. Huang,
  ``A gene selection method for microarray data based on binary pso encoding
  gene-to-class sensitivity information,'' \emph{IEEE/ACM Transactions on
  Computational Biology and Bioinformatics (TCBB)}, vol.~14, no.~1, pp. 85--96,
  2017.

\bibitem{mohamad2011modified}
M.~S. Mohamad, S.~Omatu, S.~Deris, and M.~Yoshioka, ``A modified binary
  particle swarm optimization for selecting the small subset of informative
  genes from gene expression data,'' \emph{IEEE Transactions on Information
  Technology in Biomedicine}, vol.~15, no.~6, pp. 813--822, 2011.

\bibitem{shen2008}
Q.~Shen, W.~Shi, and W.~Kong, ``Hybrid particle swarm optimization and tabu
  search approach for selecting genes for tumor classification using gene
  expression data,'' \emph{Comput. Biol. Chem.}, vol.~32, pp. 53--60, 2008.

\bibitem{chuang2008improved}
L.-Y. Chuang, H.-W. Chang, C.-J. Tu, and C.-H. Yang, ``Improved binary pso for
  feature selection using gene expression data,'' \emph{Computational Biology
  and Chemistry}, vol.~32, no.~1, pp. 29--38, 2008.

\bibitem{han2014novel}
F.~Han, W.~Sun, and Q.-H. Ling, ``A novel strategy for gene selection of
  microarray data based on gene-to-class sensitivity information,'' \emph{PloS
  one}, vol.~9, no.~5, p. e97530, 2014.

\bibitem{yang2008hybrid}
C.-S. Yang, L.-Y. Chuang, C.-H. Ke, and C.-H. Yang, ``A hybrid feature
  selection method for microarray classification.'' \emph{IAENG International
  Journal of Computer Science}, vol.~35, no.~3, 2008.

\bibitem{john1994irrelevant}
G.~H. John, R.~Kohavi, and K.~Pfleger, ``Irrelevant features and the subset
  selection problem,'' in \emph{Machine Learning Proceedings 1994}.\hskip 1em
  plus 0.5em minus 0.4em\relax Elsevier, 1994, pp. 121--129.

\bibitem{kumar2015feature}
S.~Kumar and S.~K. Singh, ``Feature selection and recognition of face by using
  hybrid chaotic pso-bfo and appearance-based recognition algorithms,''
  \emph{International Journal of Natural Computing Research (IJNCR)}, vol.~5,
  no.~3, pp. 26--53, 2015.

\bibitem{agarwal2017frbpso}
S.~Agarwal, R.~Rajesh, and P.~Ranjan, ``Frbpso: a fuzzy rule based binary pso
  for feature selection,'' \emph{Proceedings of the National Academy of
  Sciences, India Section A: Physical Sciences}, vol.~87, no.~2, pp. 221--233,
  2017.

\bibitem{chen2014new}
K.-H. Chen, L.-F. Chen, and C.-T. Su, ``A new particle swarm feature selection
  method for classification,'' \emph{Journal of Intelligent Information
  Systems}, vol.~42, no.~3, pp. 507--530, 2014.

\bibitem{yan2016hybrid}
J.~Yan, S.~Duan, T.~Huang, and L.~Wang, ``Hybrid feature matrix construction
  and feature selection optimization-based multi-objective qpso for electronic
  nose in wound infection detection,'' \emph{Sensor Review}, vol.~36, no.~1,
  pp. 23--33, 2016.

\bibitem{zhang2015feature}
Y.~Zhang, D.~Gong, Y.~Hu, and W.~Zhang, ``Feature selection algorithm based on
  bare bones particle swarm optimization,'' \emph{Neurocomputing}, vol. 148,
  pp. 150--157, 2015.

\bibitem{shi2004particle}
Y.~Shi, ``Particle swarm optimization,'' \emph{IEEE connections}, vol.~2,
  no.~1, pp. 8--13, 2004.

\bibitem{rini2011particle}
D.~P. Rini, S.~M. Shamsuddin, and S.~S. Yuhaniz, ``Particle swarm optimization:
  technique, system and challenges,'' \emph{International journal of computer
  applications}, vol.~14, no.~1, pp. 19--26, 2011.

\bibitem{hasanzadeh2013adaptive}
M.~Hasanzadeh, M.~R. Meybodi, and M.~M. Ebadzadeh, ``Adaptive cooperative
  particle swarm optimizer,'' \emph{Applied Intelligence}, vol.~39, no.~2, pp.
  397--420, 2013.

\bibitem{babu2014face}
S.~H. Babu, S.~A. Birajdhar, and S.~Tambad, ``Face recognition using entropy
  based face segregation as a pre-processing technique and conservative bpso
  based feature selection,'' in \emph{Proceedings of the 2014 Indian Conference
  on Computer Vision Graphics and Image Processing}.\hskip 1em plus 0.5em minus
  0.4em\relax ACM, 2014, p.~46.

\bibitem{mukhopadhyay2010cooperating}
S.~Mukhopadhyay and S.~Banerjee, ``Cooperating swarms: A paradigm for
  collective intelligence and its application in finance,'' \emph{International
  Journal of Computer Applications}, vol.~6, no.~10, pp. 31--41, 2010.

\bibitem{lim2014particle}
W.~H. Lim and N.~A.~M. Isa, ``Particle swarm optimization with increasing
  topology connectivity,'' \emph{Engineering Applications of Artificial
  Intelligence}, vol.~27, pp. 80--102, 2014.

\bibitem{lin2008particle}
S.-W. Lin, K.-C. Ying, S.-C. Chen, and Z.-J. Lee, ``Particle swarm optimization
  for parameter determination and feature selection of support vector
  machines,'' \emph{Expert systems with applications}, vol.~35, no.~4, pp.
  1817--1824, 2008.

\bibitem{ghamisi2015feature}
P.~Ghamisi and J.~A. Benediktsson, ``Feature selection based on hybridization
  of genetic algorithm and particle swarm optimization,'' \emph{IEEE Geoscience
  and Remote Sensing Letters}, vol.~12, no.~2, pp. 309--313, 2015.

\bibitem{kabir2012new}
M.~M. Kabir, M.~Shahjahan, and K.~Murase, ``A new hybrid ant colony
  optimization algorithm for feature selection,'' \emph{Expert Systems with
  Applications}, vol.~39, no.~3, pp. 3747--3763, 2012.

\bibitem{khushaba2008combined}
R.~N. Khushaba, A.~Al-Ani, A.~AlSukker, and A.~Al-Jumaily, ``A combined ant
  colony and differential evolution feature selection algorithm,'' in
  \emph{International Conference on Ant Colony Optimization and Swarm
  Intelligence}.\hskip 1em plus 0.5em minus 0.4em\relax Springer, 2008, pp.
  1--12.

\bibitem{bansal2011inertia}
J.~C. Bansal, P.~Singh, M.~Saraswat, A.~Verma, S.~S. Jadon, and A.~Abraham,
  ``Inertia weight strategies in particle swarm optimization,'' in \emph{Nature
  and Biologically Inspired Computing (NaBIC), 2011 Third World Congress
  on}.\hskip 1em plus 0.5em minus 0.4em\relax IEEE, 2011, pp. 633--640.

\bibitem{engelbrecht2012particle}
A.~Engelbrecht, ``Particle swarm optimization: Velocity initialization,'' in
  \emph{Evolutionary Computation (CEC), 2012 IEEE Congress on}.\hskip 1em plus
  0.5em minus 0.4em\relax IEEE, 2012, pp. 1--8.

\bibitem{coello2007evolutionary}
C.~A.~C. Coello, G.~B. Lamont, D.~A. Van~Veldhuizen \emph{et~al.},
  \emph{Evolutionary algorithms for solving multi-objective problems}.\hskip
  1em plus 0.5em minus 0.4em\relax Springer, 2007, vol.~5.

\bibitem{ray2002constrained}
T.~Ray, ``Constrained robust optimal design using a multiobjective evolutionary
  algorithm,'' in \emph{wcci}.\hskip 1em plus 0.5em minus 0.4em\relax IEEE,
  2002, pp. 419--424.

\bibitem{mostaghim2003strategies}
S.~Mostaghim and J.~Teich, ``Strategies for finding good local guides in
  multi-objective particle swarm optimization (mopso),'' in \emph{Swarm
  Intelligence Symposium, 2003. SIS'03. Proceedings of the 2003 IEEE}.\hskip
  1em plus 0.5em minus 0.4em\relax IEEE, 2003, pp. 26--33.

\bibitem{fieldsend2003using}
J.~E. Fieldsend, R.~M. Everson, and S.~Singh, ``Using unconstrained elite
  archives for multi-objective optimisation,'' 2003.

\bibitem{deb2000fast}
K.~Deb, S.~Agrawal, A.~Pratap, and T.~Meyarivan, ``A fast elitist non-dominated
  sorting genetic algorithm for multi-objective optimization: Nsga-ii,'' in
  \emph{International Conference on Parallel Problem Solving From
  Nature}.\hskip 1em plus 0.5em minus 0.4em\relax Springer, 2000, pp. 849--858.

\bibitem{leung2014new}
M.-F. Leung, S.-C. Ng, C.-C. Cheung, and A.~K. Lui, ``A new strategy for
  finding good local guides in mopso,'' in \emph{Evolutionary Computation
  (CEC), 2014 IEEE Congress on}.\hskip 1em plus 0.5em minus 0.4em\relax IEEE,
  2014, pp. 1990--1997.

\bibitem{chakraborty2011convergence}
P.~Chakraborty, S.~Das, G.~G. Roy, and A.~Abraham, ``On convergence of the
  multi-objective particle swarm optimizers,'' \emph{Information Sciences},
  vol. 181, no.~8, pp. 1411--1425, 2011.

\bibitem{lichman2013uci}
M.~Lichman \emph{et~al.}, ``Uci machine learning repository,'' 2013.

\bibitem{nguyen2018evolutionary}
B.~H. Nguyen, ``Evolutionary computation for feature selection in
  classification,'' 2018.

\bibitem{Armstrong2001}
\BIBentryALTinterwordspacing
S.~A. Armstrong, J.~E. Staunton, L.~B. Silverman, R.~Pieters, M.~L. den Boer,
  M.~D. Minden, S.~E. Sallan, E.~S. Lander, T.~R. Golub, and S.~J. Korsmeyer,
  ``Mll translocations specify a distinct gene expression profile that
  distinguishes a unique leukemia,'' \emph{Nature Genetics}, vol.~30, pp. 41 EP
  --, Dec 2001, article. [Online]. Available:
  \url{http://dx.doi.org/10.1038/ng765}
\BIBentrySTDinterwordspacing

\bibitem{SINGH2002203}
\BIBentryALTinterwordspacing
D.~Singh, P.~G. Febbo, K.~Ross, D.~G. Jackson, J.~Manola, C.~Ladd, P.~Tamayo,
  A.~A. Renshaw, A.~V. D'Amico, J.~P. Richie, E.~S. Lander, M.~Loda, P.~W.
  Kantoff, T.~R. Golub, and W.~R. Sellers, ``Gene expression correlates of
  clinical prostate cancer behavior,'' \emph{Cancer Cell}, vol.~1, no.~2, pp.
  203 -- 209, 2002. [Online]. Available:
  \url{http://www.sciencedirect.com/science/article/pii/S1535610802000302}
\BIBentrySTDinterwordspacing

\bibitem{Shipp2002}
\BIBentryALTinterwordspacing
M.~A. Shipp, K.~N. Ross, P.~Tamayo, A.~P. Weng, J.~L. Kutok, R.~C.~T. Aguiar,
  M.~Gaasenbeek, M.~Angelo, M.~Reich, G.~S. Pinkus, T.~S. Ray, M.~A. Koval,
  K.~W. Last, A.~Norton, T.~A. Lister, J.~Mesirov, D.~S. Neuberg, E.~S. Lander,
  J.~C. Aster, and T.~R. Golub, ``Diffuse large b-cell lymphoma outcome
  prediction by gene-expression profiling and supervised machine learning,''
  \emph{Nature Medicine}, vol.~8, pp. 68 EP --, Jan 2002, article. [Online].
  Available: \url{http://dx.doi.org/10.1038/nm0102-68}
\BIBentrySTDinterwordspacing

\bibitem{diaz2006gene}
R.~D{\'\i}az-Uriarte and S.~A. De~Andres, ``Gene selection and classification
  of microarray data using random forest,'' \emph{BMC bioinformatics}, vol.~7,
  no.~1, p.~3, 2006.

\bibitem{tran2016investigation}
B.~Tran, B.~Xue, M.~Zhang, and S.~Nguyen, ``Investigation on particle swarm
  optimisation for feature selection on high-dimensional data: Local search and
  selection bias,'' \emph{Connection Science}, vol.~28, no.~3, pp. 270--294,
  2016.

\end{thebibliography}

\end{document}